\documentclass{article}

\usepackage{spconf,amsmath,graphicx}
\usepackage{amsfonts,amssymb}
\usepackage{caption}
\usepackage{booktabs}
\usepackage{multirow}
\usepackage[export]{adjustbox}
\usepackage[pagebackref,breaklinks,colorlinks]{hyperref}

\usepackage{fancyhdr}

\fancypagestyle{firstpage}{
  \fancyhf{}
  \fancyhead[L]{\textit{Proceedings of IC-NIDC 2023}}
  
}

\title{Vision-language assisted attribute learning}

\name{\begin{tabular}{c}
Kongming Liang$^{1}$,
Xinran Wang$^{1}$, 
Rui Wang$^{2}$, 
Donghui Gao$^{2}$,
Ling Jin$^{2}$,
Weidong Liu$^{2}$, \\
Xiatian Zhu$^{3}$, 
Zhanyu Ma$^{1}$*\thanks{*
indicates the corresponding author.}, 
Jun Guo$^{1}$\end{tabular}}

\address{
$^{1}$School of Artificial Intelligence, Beijing University of Posts and Telecommunications\\
$^{2}$China Mobile Research Institute \quad\quad $^3$University of Surrey, UK
}
\begin{document}
%

\maketitle

\begin{abstract}
Attribute labeling at large scale is typically incomplete and partial, posing significant challenges to model optimization. Existing attribute learning methods often treat the missing labels as negative or simply ignore them all during training, either of which could hamper the model performance to a great extent.
To overcome these limitations, in this paper
we leverage the available vision-language knowledge to explicitly disclose the missing labels for enhancing model learning.
Given an image, we predict the likelihood of each missing attribute label assisted by an off-the-shelf vision-language model,
and randomly select to ignore those with high scores in training.
Our strategy strikes a good balance between fully ignoring and negatifying the missing labels, as these high scores are found to be informative on revealing label ambiguity.
Extensive experiments show that our proposed vision-language assisted loss can achieve state-of-the-art performance on the newly cleaned VAW dataset. 
Qualitative evaluation demonstrates the ability of the proposed method in predicting more complete attributes.


\end{abstract}
\section{Introduction}
\label{sec:intro}
\begin{figure}[t]
\begin{center}
\includegraphics[width=0.86\linewidth]{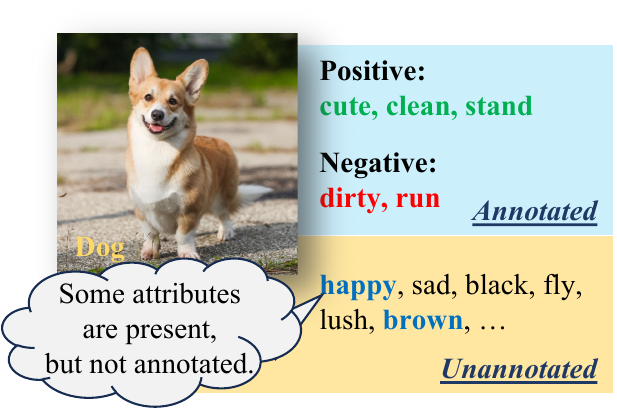}
\caption{
Illustration of typical partial attribute labeling.
} \label{fig:partial-label-diagram}
\end{center}
\vspace{-0.5cm}
\end{figure}

Visual attributes (e.g., color, shape, texture, part)~\cite{farhadi2009describing,patterson2016coco,liang-2019pami} are an important means of describing the fine-grained differences between objects. Learning to predict attributes 
underpins
many downstream applications such as image captioning~\cite{fang2015captions}, 
visual question answering (VQA)~\cite{antol2015vqa}, 
image generation~\cite{libing1,libing2}, and compositional zero-shot learning~\cite{2018-ECCV-nagarajan,mancini2021open}.
Since the space of visual attributes is often large, it is exhaustive for the annotators to label out all the  attributes per image in practice. 
For example, Pham et al. \cite{pham2021learning} constructed a large-scale attribute dataset (VAW), with 2260 objects and 620 attributes. 
On average, only 3.56 attributes are annotated for each image, 
largely incomplete and partial (see Fig.~\ref{fig:partial-label-diagram}).
This problem poses significant challenges to model optimization, which however is still understudied in the literature~\cite{2017-IJCAI-liang}.

\begin{figure*}[t]
\begin{center}
\includegraphics[width=0.9\linewidth]{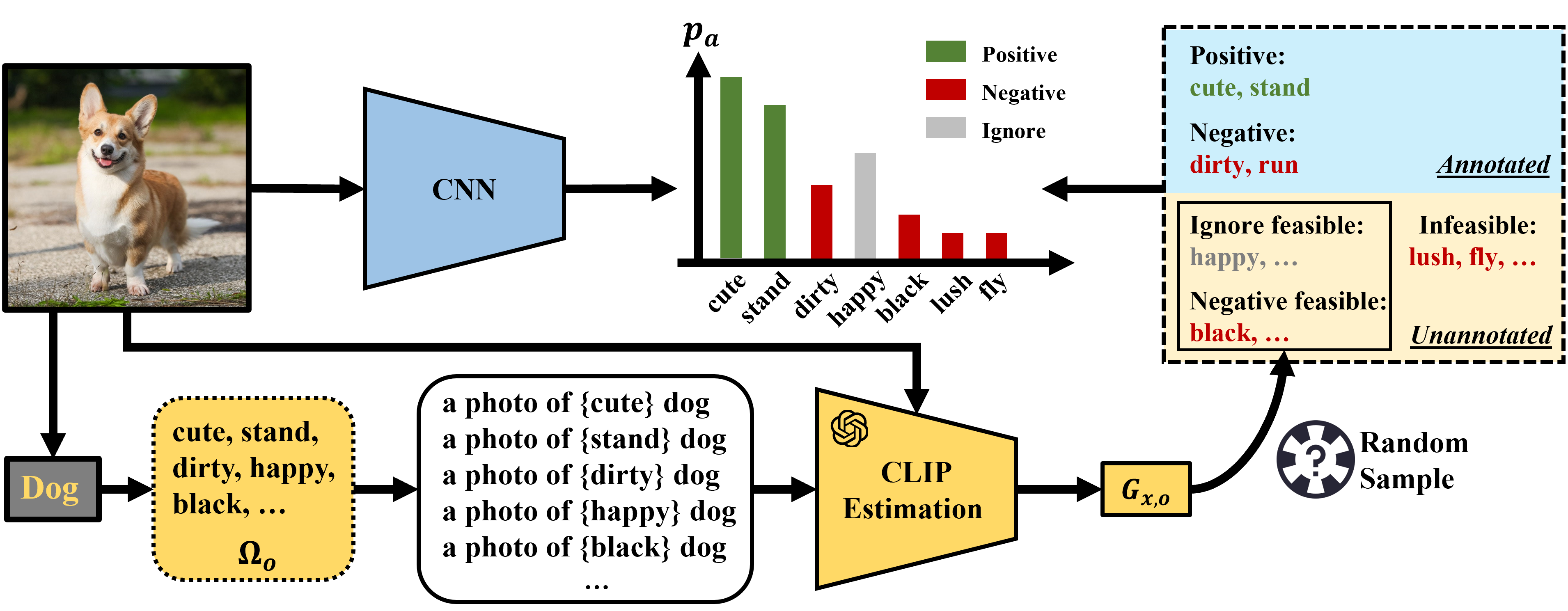}
\caption{Overview of the proposed vision-language assisted attribute learning. 
Given a training image with partial labeling, we exploit a vision-language model to estimate the presence probability of each attribute ($\Omega_o$).
As this prediction could be largely unreliable, we treat those attributes with high probabilities as ambiguity.
For model optimization, we randomly ignore some of the ambiguous attributes 
and negatify the rest missing ones.
}
\label{fig:method-diagram}
\end{center}
\vspace{-0.4cm}
\end{figure*}

Typically, the number of missing/unannotated attribute labels 
is even more than that of labeled ones.
Thus, dealing with the missing attributes is critical.
There are three existing representative strategies.
The first is discarding the loss terms with missing attributes,
termed as \texttt{ignoring} \cite{Durand_2019_CVPR}.
As only a small portion of labels are used for model training, the performance is naturally limited.
The second simply treats all the missing labels as negative,
termed as \texttt{negatifying} \cite{kundu2020exploiting}.
This however would introduce a plenty of false labels
to those missing attributes.
The third aims to combine the above two strategies by \texttt{selecting} one for each attribute \cite{Ben-Baruch_2022_CVPR,liang2023attribute}.
Nonetheless, this method falls into {\em the chicken or the egg causality dilemma} given that predicting the attributes and training the attribute model (the two key components) are mutually depending on each other.


To overcome all the aforementioned limitations, we propose to leverage the available vision-language knowledge for enhancing attribute learning under partial labeling.
Instead of struggling to predict the missing labels accurately, we set up to find out what attributes are ambiguous and thus need to be ignored during training.
This is because, achieving this objective is less demanding and likely more achievable, without the notorious dilemma as suffered by \cite{Ben-Baruch_2022_CVPR}.
Specifically, we first estimate all the missing attributes of a given training image assisted by an off-the-shelf vision-language model (e.g., CLIP \cite{2021-CLIP}).
Considering that such predictions serve as an indicator of the presence ambiguity,
we take those with high probability scores as {\em ambiguous attributes}.
To further tackle the predicting uncertainty,
we randomly select some of these ambiguous attributes
and ignore their loss contributions during training,
whilst negatifying the remaining missing attributes.
This strategy could yield a better trade-off between the ignoring and negatifying strategies.
Experimental results demonstrate that the proposed vision-language assisted attribute learning method achieves the new state-of-the-art on the standard large attribute recognition benchmark.


\section{Method}
\label{sec:method}

In this section, we start with the formalization of partial attribute learning. Then we leverage an off-the-shelf vision-language model to estimate the presence probability for each unannotated attribute. Finally, we design a selective loss by randomly ignoring the loss of the ambiguous attributes. The overall method is summarized in Fig.~\ref{fig:method-diagram}.

\subsection{Attribute Learning}
Attribute learning aims to predict the presence or absence of each pre-defined attribute for an input visual instance. Considering the limited annotation cost, the majority of the attributes are left unannotated which may hinder the learning process. For the visual instance $x$, we denote its corresponding object label as $o$ and attribute vector as $y = \{y_a\}^A_{a=1}$, where $A$ denotes the number of attributes. If the attribute $a$ is present, absent or unannotated in the visual instance, $y_a$ is represented as 1, -1, 0 respectively. Then the positive and negative attribute sets of $x$ are constructed as $\mathcal{P}_{x} = \{a|y_a = 1\}$ and $\mathcal{N}_x = \{a|y_a = -1\}$. And the unannotated attribute set is denoted as $\mathcal{U}_x = \{a|y_a = 0\}$. 

For an image $x$ with its object label $o$, an attribute classifier $f_a$ is trained to predict whether $x$ possesses the attribute $a$ or not. The prediction probability is denoted as $p_a=f_a(x,o; \theta_a)$. Then the general form of attribute learning loss can be formulated as follows,
\begin{equation}
\mathcal{L}(x,o) = \sum_{a \in \mathcal{P}_x} \mathcal{L}^+_a (x,o)
+ \sum_{a \in \mathcal{N}_x} \mathcal{L}^-_a (x,o)
+ \sum_{a \in \mathcal{U}_x} \mathcal{L}^u_a (x,o),
\end{equation}
where $\mathcal{L}^+_a (x,o)$, $\mathcal{L}^-_a (x,o)$ and $\mathcal{L}^u_a (x,o)$ are the loss terms calculated by positive, negative, and unannotated attributes. In this paper, we use the asymmetric loss (ASL)~\cite{2021-CVPR-ASL} to mitigate the negative-positive imbalance problem. 
We denote the positive loss term as $\mathcal{L}^+_a (x,o) = (1-p_a)^{\gamma^+} \text{log}(p_a)$ and the negative loss term as $\mathcal{L}^-_a (x,o) = (p_a)^{\gamma^-} \text{log}(1 - p_a)$. ASL can focus on the hard samples dynamically while controlling the contribution propagated from the positive and negative samples at the same time. For positive and negative loss, we use $\gamma^+$ and $\gamma^-$ to denote their focusing parameters respectively.

\subsection{Estimating the Unannotated Attributes}
For the unannotated attribute labels, previous methods may \texttt{ignore} their loss by setting the $\mathcal{L}^u_a (x) = 0$ or treat them as \texttt{negative} by setting the $\mathcal{L}^u_a (x) = log(1 - p_a)$. However, both methods above are inefficient in model training since \texttt{ignoreing} mode only includes a portion of the labels, and \texttt{negatifying} mode introduces extra false labels as some missing attributes may be present. 
Another solution combines the above two strategies by selecting one for each attribute. Nonetheless, this method falls into the chicken or the egg causality dilemma as predicting the attributes and training the attribute model are mutually dependent on each other.
In this paper, we use CLIP \cite{2021-CLIP}, a powerful vision-language model, to estimate the probability of each attribute being present in a given object instance.
Specifically, we first calculate the object-attribute co-occurrence on the training data. Then, if there is any sample belonging to the object $o$ and annotated with attribute $a$ simultaneously, $a$ will be appended to the feasible set $\Omega_o$. Only the attributes contained in the feasible set are further processed by CLIP.

To use CLIP for unannotated attribute estimation, the object and attribute names are transformed into natural language prompts such as \texttt{A photo of [attribute][object]}. The prompts are further processed to get the embeddings for each word. Then, all the embeddings are passed through the text encoder to obtain the text representations. For the visual counterpart, we can extract the visual representations based on the image encoder. Given a visual instance $x$, we denote its visual representations from CLIP as $x_v$. And the text representations for attribute $a$ and object $o$ can be denoted as $t_{a,o}$.
Then we compute the cosine similarities between the visual representations and the text representations to compute the presence probability of the attribute $a$:
\begin{equation}
g(a|x,o)=\frac{\text{exp}(x_v \cdot t_{a,o}/\tau)}{\sum_{\hat{a} \in \Omega_o} \text{exp}(x_v \cdot t_{\hat{a},o}/\tau)},
\end{equation}
where $\tau$ is a temperature parameter and only the feasible attributes of the object $o$ are included for probability estimation. The likelihood vector of feasible attributes can be denoted as $\mathbf{G}_{x,o} = \{g(a|x,o) \mid \forall a \in \Omega_o\}$.

\begin{figure}[t]
\centering
\centerline{\includegraphics[width=9.0cm]{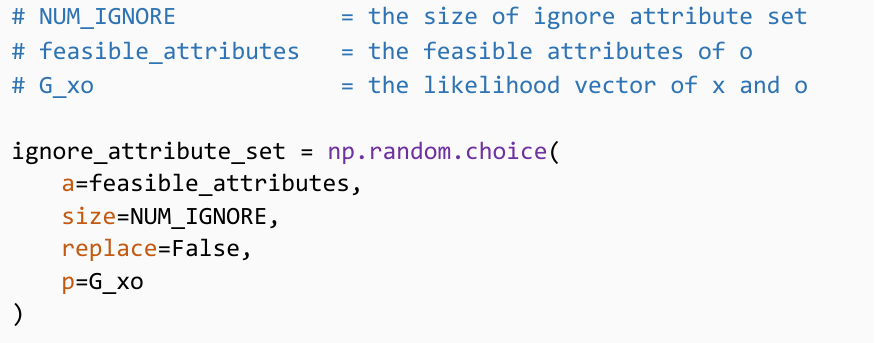}}
\caption{Numpy-like pseudocode for the implementation of ignored attribute sampling.}
\label{fig:code}
\end{figure}

\subsection{Vision-language Guided Selective Loss}
For partially labeled attribute learning, the number of unannotated attributes is much more than the number of positive and negative attributes (e.g. $|\mathcal{P}_x \cup \mathcal{N}_x| \ll |\mathcal{U}_x|$). Therefore, how to process the unannotated attributes is of great importance. With the aid of attribute likelihood provided by CLIP, we can estimate the probability of an unannotated attribute being present in a visual instance. The attributes with high probability scores are considered as ambiguous attributes. To further tackle the predicting uncertainty, we randomly select some of these ambiguous attributes and ignore their loss contributions during training, whilst negatifying the remaining missing attributes. 
We formalize the above vision-language guided selective loss as follows,
\begin{equation}
\label{eqn:unannotated_loss}
\mathcal{L}^u_a (x,o) = 
\begin{cases}
 \omega_a (p_a)^{\gamma^{\oplus}} \text{log}(1 - p_a) & \text{ if } a \in \Omega_o \\
 {(p_a)^{\gamma^{\ominus}}} \text{log}(1 - p_a)  & \text{ if } a \notin \Omega_o
\end{cases}
,
\end{equation}
where $\gamma^\oplus$ and $\gamma^\ominus$ denote the feasible and infeasible focusing parameters of the object $o$ respectively. We set $\gamma^\oplus < \gamma^\ominus$ to decay the feasible term with a lower rate than the infeasible one since most infeasible attributes are easy negative samples. $\omega_a$ is used to ignore the suspected positive attributes. We generate random samples from the non-uniform distribution $\mathbf{G}_{x,o}(a)$ of feasible attributes to construct the ignore attribute set $\Omega_{Ignore}$. 
In Fig.\ref{fig:code}, we include pseudocode for the core implementation of ignored attribute sampling. The parameter $\omega_a$ in Eqn.~(\ref{eqn:unannotated_loss}) can be defined as follows,

\begin{equation}
\omega_a = 
\begin{cases}
  0 & \text{ if } a \in \Omega_{Ignore} \\
  1  & \text{ if } a \notin \Omega_{Ignore}
\end{cases}
.
\end{equation}
The loss contributed by the ignored attribute is discarded during training.

\section{Experiment}
\label{sec:experiment}
In this section, we first present the dataset and evaluation metrics for large-scale attribute learning.
Next, we compare the proposed method with existing state-of-the-arts. Finally, we conduct ablation studies to show the hyperparameter effect and demonstrate the qualitative results.

\begin{table*}
\caption{Comparison with the state-of-the-art methods of partially labeled multi-label learning on the cleaned VAW dataset. \texttt{ignoring} means discarding the loss of all the unannotated labels. And \texttt{negatifying} means considering all the unannotated labels to be negative.}
\footnotesize
\label{tab:result-vaw2}
\begin{tabular}{l|c|ccc|cccccccc}
\hline
\multicolumn{1}{c|}{\multirow{2}{*}{Methods}} & \multirow{2}{*}{\begin{tabular}[c]{@{}c@{}}Overall \\ (mAP)\end{tabular}} & \multicolumn{3}{c|}{Class imbalance (mAP)}        & \multicolumn{8}{c}{Attribute types (mAP)}                                                                                             \\
\multicolumn{1}{c|}{}                         &                                                                           & Head           & Medium         & Tail           & Color          & Material       & Shape          & Size           & Texture        & Action         & State          & Others         \\ \hline
CE-\texttt{ignoring}                                    & 61.18                                                                     & 68.19          & 58.31          & 44.88 & 52.99          & 56.90          & 61.87          & 59.69          & 61.61          & 53.49          & 58.49          & \textbf{65.95} \\
CE-\texttt{negatifying}~\cite{patterson2016coco}                                  & 61.37                                                                     & 69.61          & 58.41          & 40.91          & 60.82          & \textbf{66.42} & 61.03          & 58.60          & 62.42          & 53.43          & 57.60          & 62.87          \\
ASL-\texttt{ignoring}                                   & 61.29                                                                     & 68.45          & 58.49          & 44.18          & 52.76          & 57.94          & 61.92          & 59.51          & 62.39          & 53.88          & 58.45          & 65.93          \\
ASL-\texttt{negatifying}~\cite{2021-CVPR-ASL}                                 & 62.01                                                                     & 70.02          & 58.64          & 43.61          & 61.96          & 65.42          & 62.44          & \textbf{60.91} & 62.33          & 52.73          & 58.18          & 63.77          \\
PSL~\cite{Ben-Baruch_2022_CVPR}                                           & 62.28                                                                     & 69.83          & 59.47          & 43.82          & 57.80          & 64.91          & 62.74          & 60.35          & 63.55 & 52.87          & 58.57          & 65.36          \\ \hline
Ours                                          & \textbf{64.06}                                                            & \textbf{71.38} & \textbf{61.67} & \textbf{45.16} & \textbf{63.87} & 66.36          & \textbf{63.55} & 60.46          & \textbf{64.18}          & \textbf{57.30} & \textbf{61.01} & 65.73          \\ \hline
\end{tabular}
\end{table*}

\begin{figure*}
    
    \begin{minipage}{0.65\linewidth}
    
    \includegraphics[width=\linewidth]{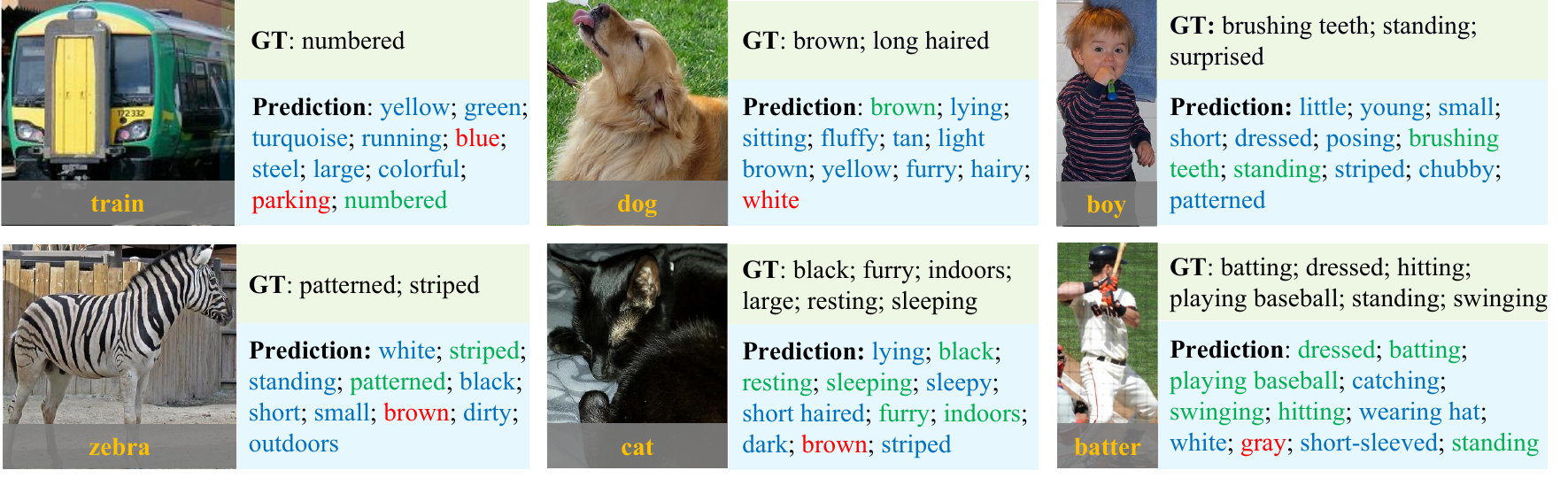}
    
    \caption{Qualitative results. We show the ground truth positive attribute labels and the Top-10 attributes predicted by the proposed method. The correct predictions are highlighted in green, the incorrect predictions are highlighted in red, and the correct predictions without annotation are highlighted in blue. }
    \label{fig:qualitative}

    \end{minipage}
  \hfill
  \begin{minipage}{0.34\linewidth}
    \centering
    \captionof{table}{The ablation study of the parameters in the vision-language guided selective loss. The results are shown for different values of NUM\_IGNORE.}
    \label{tab:ablation}
    \begin{tabular}{c|c}
        \hline
        NUM\_IGNORE & Overall (mAP) \\ \hline
        10                & 63.70         \\
        20                & 63.96         \\
        30                & \textbf{64.06}\\
        40                & 63.81         \\
        50                & 63.83         \\
        ALL               & 59.41         \\ \hline
        \end{tabular}
  \end{minipage}
\end{figure*}

\subsection{Dataset and Evaluation Metric}

We use the currently released VAW dataset~\cite{Ben-Baruch_2022_CVPR} in the experiments. To overcome the negative impact of tiny objects, we discard the object instances that are smaller than $50 \times 50$. Then we merge the objects and attributes based on their plurals and synonyms (e.g. \textit{plane}, \textit{airplane}, and \textit{airplanes}). The cleaned version of VAW contains 170,407 visual instances, 1763 object categories, and 591 attributes.
For evaluation, we follow the previous multi-label learning and attribute prediction methods 
to report the mAP for all attributes. Since the dataset is partially labeled, only the annotated attributes are included for evaluation~\cite{pham2021learning}.

\subsection{Implementation Details}
ResNet-50 is adopted as the backbone for attribute prediction. During training, we select Adam as the optimizer and set the initial learning rate as $10^{-5}$ for the pre-trained parameters and $7\times 10^{-4}$ for the newly added parameters. The model is optimized for 12 epochs with a batch size of 64. We set $\gamma^+=1, \gamma^-=2, \gamma^\oplus=4, \gamma^\ominus=7$. For the unannotated label estimation, ViT-B/32 is set as the vision transformer of CLIP.

\subsection{Comparison with State-of-the-art Methods}
The proposed method is compared with multi-label cross entropy (CE)~~\cite{patterson2016coco}, asymmetric loss (ASL)~\cite{2021-CVPR-ASL}, and Partial-ASL (PSL)~\cite{Ben-Baruch_2022_CVPR} which are the state-of-the-art methods of partially labeled multi-label classification. We use the same backbone ResNet-50 for the above methods to make a fair comparison. Both \textit{ignore} and \textit{negative} modes for CE and ASL are reported in our experiments. 
The overall experiments are shown in Table~\ref{tab:result-vaw2}. Compared to the other methods, our proposed vision-language guided selective loss achieves 64.06 mAP score and outperforms the state-of-the-art method (PSL).
To consider the positive and negative label imbalance, we report the mAP scores of head, medium, and tail attributes. Similarly, our method can achieve better performance under different imbalance conditions.
In addition, our proposed loss outperforms the other methods in terms of most of the attribute types. The above experiments consistently demonstrate the effectiveness of the proposed vision-language assisted attribute estimation procedure.


\subsection{Ablation Study and Qualitative Results}
The effectiveness of the number of ignored attributes is investigated on partially labeled attribute learning. As demonstrated in Table~\ref{tab:ablation}, the performance firstly increases as more attributes are involved in ignore set. Then the performance keeps stabilizing when the size of ignore set is larger than 30. If we regard all the attributes to be ignored, the overall performance decreases to 59.41, which is even lower than the baseline methods. We show some qualitative results in Fig.~\ref{fig:qualitative}. The results show the proposed method can predict more complete attributes, though only part of them are manually labeled.


\section{Conclusion}
\label{sec:conclusion}

A vision-language assisted selective loss is proposed for partially labeled attribute learning. The proposed method can well estimate the presence probability of unannotated attributes by an off-the-shelf vision-language model and further randomly select to ignore those with high scores in training. The proposed loss achieves the best performance on the newly cleaned VAW dataset. The qualitative results demonstrate the proposed vision-language assisted selective loss can predict the attributes more completely.



\section{Acknowledgements}
\label{sec:prior}
\vspace{-0.05cm}

This work is supported in part by National Natural Science Foundation of China (NSFC) No. 62106022, 62225601, U19B2036, and in part by Beijing University of Posts and Telecommunications-China Mobile Research Institute Joint Innovation Center, and in part by Program for Youth Innovative Research Team of BUPT.
\vfill\pagebreak

\bibliographystyle{IEEEbib}
\small
\bibliography{refs}

\end{document}